\documentclass{article}
\usepackage{ijcai26}

\usepackage{times}
\usepackage{soul}
\usepackage{url}
\usepackage[hidelinks]{hyperref}
\usepackage[utf8]{inputenc}
\usepackage[small]{caption}
\usepackage{graphicx}
\usepackage{amsmath}
\usepackage{amsthm}
\usepackage{booktabs}
\usepackage{algorithm}
\usepackage{algorithmic}

\usepackage{newfloat}
\usepackage{listings}
\usepackage{amssymb}
\usepackage{multirow} 
\usepackage{array} 
\usepackage{graphicx} 
\usepackage{makecell} 
\usepackage{mathtools}
\usepackage{colortbl}
\usepackage{pifont}
\usepackage{arydshln}
\usepackage{bm}
\usepackage{times}
\usepackage{helvet}
\usepackage{courier}
\usepackage[dvipsnames]{xcolor}

\newcommand{\redup}[1]{${\color{RedOrange}\uparrow #1}$}

\title{SPHENIC: Topology-Aware Multi-View Clustering for Spatial Transcriptomics}

\author{
Chenkai Guo$^{2,5}$
\and
Yikai Zhu$^1$\and
Renxiang Guan$^{3}$\and
Jinli Ma$^1$\and
Siwei Wang$^4$\and\\
Ke Liang$^3$\and
Guangdun Peng$^{2,5}$\And
Dayu Hu$^{1,\dag}$\\
\affiliations
$^1$College of Medicine and Biological Information Engineering, Northeastern University\\
$^2$Guangzhou Institutes of Biomedicine and Health, Chinese Academy of Sciences\\
$^3$College of Computer Science and Technology, National University of Defense Technology\\
$^4$Intelligent Game and Decision Lab, Academy of Military Sciences\\
$^5$University of Chinese Academy of Sciences\\
$^{\dag}$ Corresponding Author\\
\emails
shaunak.guo@gmail.com,
hudy@bmie.neu.edu.cn
}

\usepackage{pifont}  
\usepackage{arydshln} 
\makeatletter
\newcommand{\thickhline}{%
    \noalign {\ifnum 0=`}\fi \hrule height 1pt
    \futurelet \reserved@a \@xhline
}
\makeatother
\begin{document}

\maketitle

\begin{abstract}
Spatial transcriptomics clustering is pivotal for identifying cell subpopulations by leveraging spatial location information. While recent graph-based methods modeling cell–cell interactions have improved clustering accuracy, they remain limited in two key aspects: (i) reliance on local aggregation in static graphs often fails to capture robust global topological structures (e.g., loops and voids) and is vulnerable to noisy edges; and (ii) dimensionality reduction techniques frequently neglect spatial coherence, causing physically adjacent spots to be erroneously separated in the latent space. To overcome these challenges, we propose SPHENIC, a \textbf{S}patial \textbf{P}ersistent \textbf{H}omology–\textbf{E}nhanced \textbf{N}eighborhood \textbf{I}ntegrative \textbf{C}lustering method. Specifically, it explicitly incorporates topology-invariant features into the clustering network to ensure robust representation learning against noise. Furthermore, we design a dual-regularized optimization module that imposes spatial constraints alongside distributional optimization, ensuring that the embedding space preserves the physical proximity of cells. Extensive experiments on 11 benchmark datasets demonstrate that SPHENIC outperforms state-of-the-art methods by 4.19\%–9.14\%, validating its superiority in characterizing complex tissue architectures.
\end{abstract}

\section{Introduction}
Spatial transcriptomics has transformed the study of biological systems by providing spatially resolved gene-expression profiles in intact tissues \cite{rao2021exploring,moses2022museum,tian2023expanding_vista,williams2022introduction_to_ST,jain2024st_in_health_disease}. A critical step in this approach is the accurate spatial clustering required to delineate spatially coherent cellular domains in which cells exhibit both molecular similarity and physical proximity. Accurate identification of these domains may have far-reaching implications, including guiding therapeutic target discovery and advancing our understanding of complex biological systems and cellular behaviors \cite{gulati2025profiling,wang2024comprehensive_clustering_review}.

Current spatial transcriptomics clustering methods can be broadly classified into three categories: (i) traditional community-detection approaches (e.g., Seurat and Scanpy), which primarily emphasize gene expression while giving limited consideration to spatial organization \cite{wolf2018scanpy,satija2015seurat}; (ii) deep-learning–based methods, including GraphST and STAIG, which explicitly model spatial–transcriptional relationships through graph structures to identify biologically meaningful spatial domains \cite{long2023graphst,yang2025staig}; and (iii) supervised methods, such as PASSAGE, which incorporates phenotypic guidance into the clustering process \cite{guo2025passage}. The latter two categories have attracted substantial attention and are widely regarded as significant improvements over approaches that rely solely on the gene expression matrix \cite{li2025gnn_in_sc}. Among these approaches, graph-based analysis plays a pivotal role, as accurate learning of the underlying topological structure is essential for identifying cell–cell associations and, consequently, achieving high clustering accuracy.

\begin{figure}[t]
	\centering
	\includegraphics[width=0.45\textwidth]{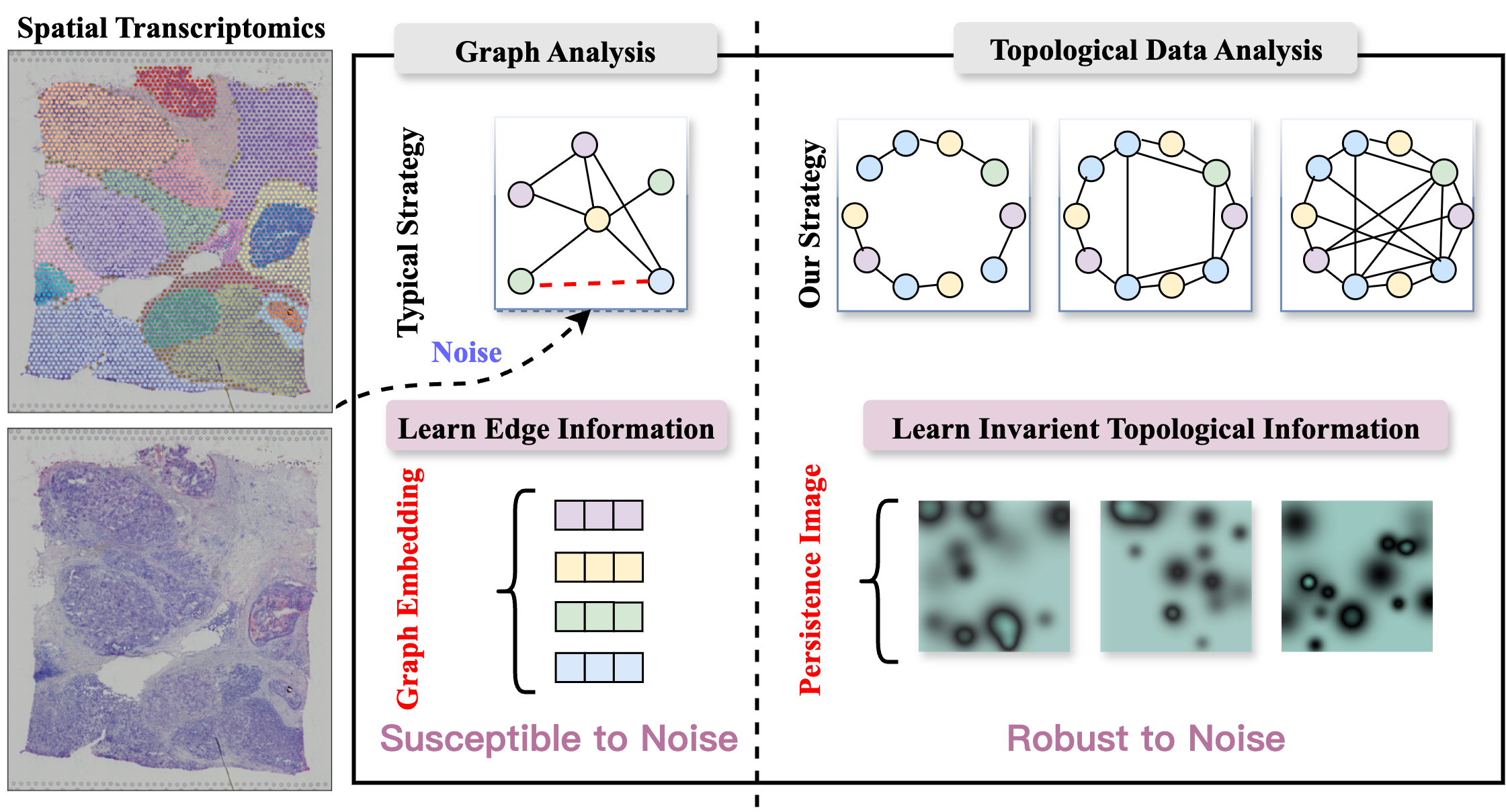} 
	\caption{Comparison of static graph learning and invariant topological feature learning.}
	\label{fig_back}
 \vspace{-5pt}
\end{figure}

Despite recent progress, existing graph-based spatial clustering approaches still face two major challenges: (i) \textbf{Susceptibility to Noise:} Most methods rely on graph representations to capture topological patterns (Fig.~\ref{fig_back}). However, such representations are often sensitive to noise, making them unreliable for preserving global geometric structures, such as loops and voids. Although some approaches incorporate persistent homology, they often fail to capture complex multidimensional structures, which hampers the learning of robust and topology-invariant representations. (ii) \textbf{Inadequate Spatial Coherence:} In addition, most existing methods model spatial relationships using simple adjacency graphs. This strategy fails to capture sufficient spatial context, leading to embeddings that distort neighborhood relationships and incorrectly separate physically adjacent spots in the latent space.

To address these limitations, we propose SPHENIC, a spatial clustering framework tailored for spatial transcriptomics data. Specifically, SPHENIC extends persistent homology to effectively learn topology-invariant features. It then constructs modality-specific views and fuses them using a multi-view Graph Convolutional Network (GCN) to learn comprehensive consensus representations. In addition, we develop a dual-regularized optimization module, which incorporates spatial constraints and distribution optimization to further refine cell distributions and embedding quality by increasing the similarity between a cell’s embedding and those of its spatial neighbors while decreasing similarity to non-neighboring cells. Extensive experiments demonstrate the superiority of our proposed SPHENIC. Our main contributions are summarized as follows:
\begin{itemize}
	\item \textit{Framework}: We pioneer the integration of extended persistent homology into spatial clustering frameworks. To the best of our knowledge, this study represents one of the earliest attempts to incorporate topology-invariant features for improving spatial clustering.
	\item  \textit{Algorithm}: We develop a novel dual-regularized optimization component that integrates spatial constraints with distributional optimization to further enhance cell distribution accuracy and embedding quality. Moreover, this component ensures that physically adjacent spots are assigned similar embeddings.
	\item \textit{Evaluation}: Extensive experiments have been conducted across three distinct spatial transcriptomics datasets, encompassing a total of eleven tissue slices, and eight current state-of-the-art baselines, demonstrating the consistent superiority of SPHENIC over existing state-of-the-art methods.	
\end{itemize}

\section{Related Work}
\subsection{Clustering in Spatial Transcriptomics}
Clustering strategies for spatial transcriptomics data can be broadly classified into three paradigms \cite{liu2024comprehensive_graph_clustering_review,zahedi2024dl_ST_review,hu2024benchmarking_ST_clustering,yuan2024benchmarking_ST_clustering_2}. (i) Traditional community-detection methods, implemented in tools such as Seurat and Scanpy, employ the Louvain or Leiden algorithm to cluster cells on the basis of gene expression similarity \cite{wolf2018scanpy,satija2015seurat,blondel2008Louvain,traag2019leiden}. Although computationally efficient, these methods typically treat spatial information as secondary and may therefore overlook the critical aspects of tissue organization. (ii) Deep-learning-based techniques have recently emerged to more effectively integrate spatial context with gene expression. For example, SpaGCN employs GCN to model spatial neighbourhoods, whereas stLearn jointly incorporates the expression profiles, spatial coordinates, and histological image features \cite{hu2021spagcn,pham2023robust,kipf2016GCN}. Spatial-MGCN further advances this paradigm by introducing multi-view graph learning framework \cite{wang2023spatial-mgcn}. (iii) Supervised methods leverage phenotypic guidance to delineate biologically meaningful spatial domains and molecular signatures. PASSAGE exemplifies this strategy by learning the phenotype-associated signatures through a two-step contrastive graph-learning framework, thereby identifying domains that correspond to the known anatomical or pathological phenotypes \cite{guo2025passage}. However, none of the existing approaches integrates invariant topological information into the spatial clustering process.

\subsection{Topological Data Analysis}
While traditional geometric deep learning outperforms in capturing local dependencies, it often struggles to quantify global structural properties invariant to deformation. Topological data analysis (TDA) leverages the rigorous framework of algebraic topology to extract multi-scale structural features, like connected components, loops, and voids, which offer a robust complement to standard graph representations \cite{pham2025TDA_in_graph_learning,wasserman2018TDA_review,su2025tda_tdl_review}. Recent progress can be summarized along three key directions: (i) Topologically-Augmented Architectures, standard Message Passing Neural Networks (MPNNs) are fundamentally limited by the Weisfeiler-Lehman hierarchy, often fail to distinguish non-isomorphic graphs with identical local neighborhoods (e.g., cycles vs. separated paths) \cite{gilmer2020message}. To bridge this gap, architectures like TopoGCL \cite{chen2024topogcl} augment graph neural networks by injecting persistent homology features into the message-passing schema, thereby enabling more robust modeling of higher-order interactions in molecular and social graphs; (ii) Topological Regularization Techniques, exemplified by TopoReg \cite{pmlr-v89-chen19g-toporeg}, introduce the filtration-based losses that preserve critical connectivity patterns within latent representations, which significantly improve performance of graph-classification task; and (iii) Topology-Guided Generation, in the generative domain, ensuring the global validity of synthesized structures remains challenging for purely probabilistic models. Therefore, methods like TopoDiff \cite{zhang2025topodiff} employ topological signatures to steer the synthesis of graphs possessing desired structural properties, a capability that is particularly valuable for drug-discovery applications. Collectively, these advances demonstrate topology's transformative potential for graph representation learning by capturing multi-scale relationships, enabling a wide range of biomedical applications \cite{hernandez2024tda_in_cardiovascular,skaf2022tda_in_biomedicine}.

\begin{figure*}[t]
	\centering
	\includegraphics[width=1\textwidth]{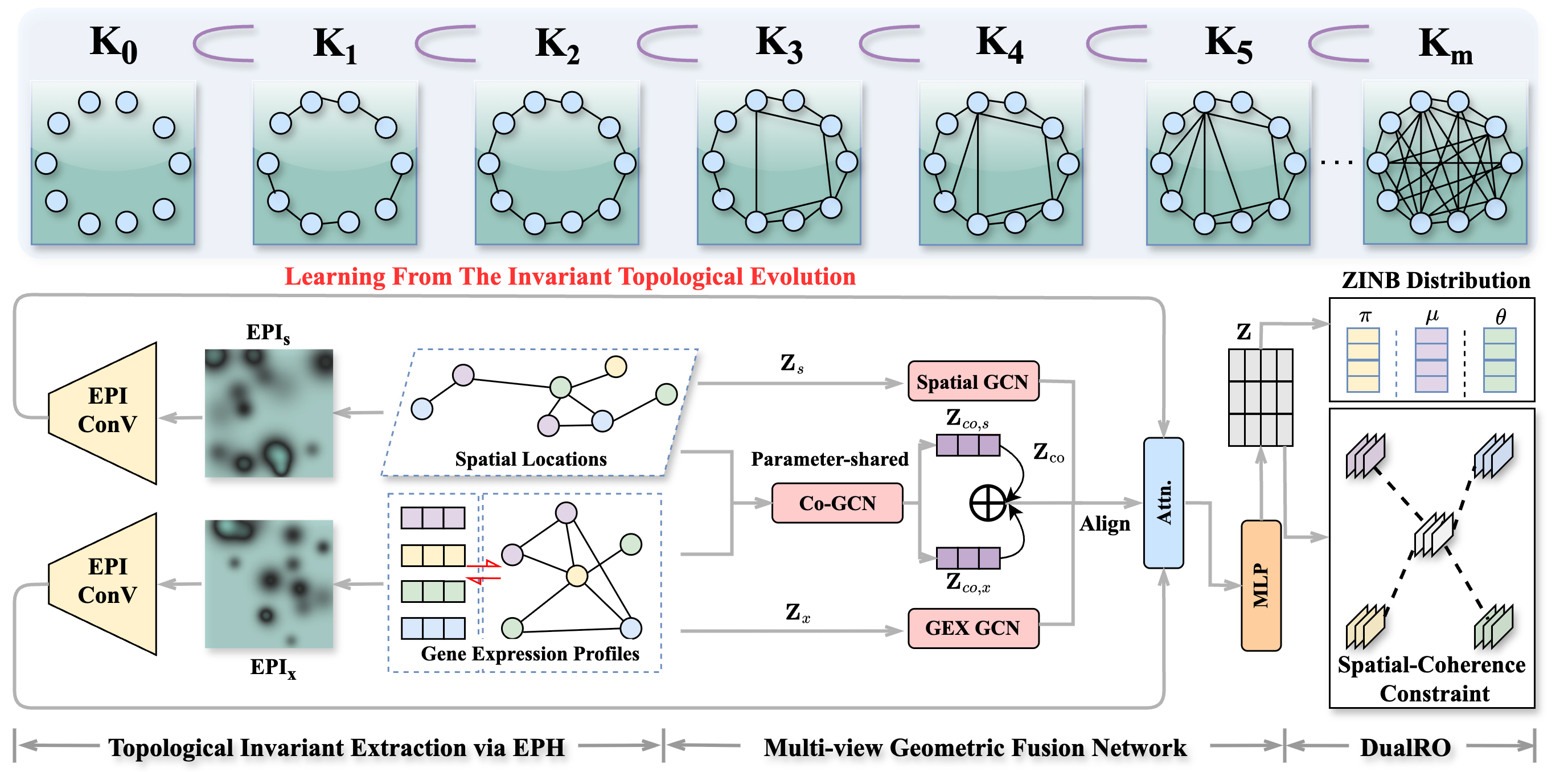} 
	\caption{The proposed SPHENIC framework. Invariant topological information is incorporated into the multi-view graph fusion network to enhance the representation of different cells by EPH feature extraction. Then, the Dual-Regularized Optimization module is designed to ensure the spatial distribution and consistency of spatial clustering results.}
	\label{fig_framework}
 \vspace{-10pt}
\end{figure*}

\section{Methods}

\subsection{Preliminaries}
\noindent\textbf{Notations.}
Formally, given a spatial transcriptomics dataset consisting of $N$ spots and $G$ genes, we denote the gene expression matrix as $\mathbf{X} \in \mathbb{R}^{N \times G}$, where each row $\mathbf{x}_i$ represents the gene expression profile of the $i$-th spot. Associated with these profiles is the spatial coordinate matrix $\mathbf{S} \in \mathbb{R}^{N \times 2}$, where $\mathbf{s}_i$ denotes the two-dimensional physical coordinates of the $i$-th spot. Additionally, we formulate the learning process over the set of modalities $\mathcal{M} = \{s, x\}$, where $s$ denotes the spatial domain and $x$ denotes the gene expression profile (GEX). To capture both transcriptomic similarity and physical proximity, we model the data as a dual-view graph system over the vertex set $\mathcal{V} = 	\{v_1, \dots, v_N\}$. We define the following graphs: 
\begin{itemize}
    \item \textbf{The Spatial Graph} $\mathcal{G}_{s} = (\mathcal{V}, \mathcal{E}_{s}, \mathbf{A}_{s})$, in which the edge set $\mathcal{E}_{s}$ and the weighted adjacency matrix $\mathbf{A}_{s}$ are constructed based on pairwise Euclidean distances in $\mathbf{S}$, thereby encoding the underlying physical tissue architecture
    \item \textbf{The Expression Graph} $\mathcal{G}_{x} = (\mathcal{V}, \mathcal{E}_{x}, \mathbf{A}_{x})$, where edges connect spots with similar transcriptomic profiles derived from $\mathbf{X}$, capturing biological functional relationships among spots.
\end{itemize}
\noindent\textbf{Problem Formulation.}
In our problem setting, the goal is to learn a mapping function
$f: (\mathbf{X}, \mathbf{S}, \mathcal{G}_s, \mathcal{G}_x) \rightarrow \mathbf{Z}$, where $\mathbf{Z} \in \mathbb{R}^{N \times d}$ represents a low-dimensional latent embedding with $d \ll G$. Ideally, the learned embedding $\mathbf{Z}$ should satisfy three key properties: (i) \textit{Topological Invariance}, preserving global geometric structures (e.g., loops and voids) in a noise-robust manner; (ii) \textit{Semantic Consistency}, aiming to learn a consensus representation that effectively integrates heterogeneous gene expression and spatial profiles; and (iii) \textit{Spatial Coherence}, ensuring that physically adjacent spots share similar embeddings, as dimensionality reduction may otherwise erroneously separate neighboring locations in the latent space. Finally, the learned embeddings $\mathbf{Z}$ are clustered to identify spatial domains.

\subsection{Topological Invariant Extraction via EPH}

To capture high-order structural invariants that are typically overlooked by graph neural networks, we employ Extended Persistent Homology (EPH). This approach transforms discrete graph data into rigorous topological descriptors through a unified analytical pipeline. The foundation of this pipeline is a filtration process, which constructs a sequence of nested subgraphs by varying a threshold parameter $\epsilon$ over each modality graph. While standard filtrations often rely solely on feature similarity, we introduce an additional geometric regularizer in the expression view to suppress long-range noise. 

Let $r$ denote a fixed spatial radius, and let $\epsilon$ represent a dynamic threshold for expression dissimilarity. At scale $\epsilon$, we define the edge set $\mathcal{E}_{x}^{\epsilon} = \{ (u,v) \mid \omega_x(u,v) \le \epsilon \text{ and } \omega_s(u,v) \le r \}$. Based on this definition, we construct a filtration sequence $\mathbb{F}_m$ for each modality $m \in \mathcal{M}$, consisting of nested subgraphs that evolve as $\epsilon$ increases.
\begin{equation}
    \mathbb{F}_m = \bigl\{ \mathcal{G}_m^{\epsilon} = (\mathcal{V}, \mathcal{E}_m^{\epsilon}) \mid \mathcal{G}_m^{\epsilon_1} \subseteq \mathcal{G}_m^{\epsilon_2} \dots \text{ for } \epsilon_1 \le \epsilon_2 \bigr\}.
\end{equation}

Upon constructing the filtration sequence, we track the evolution of topological features to identify significant structural invariants. By applying homology inference to $\mathbb{F}_x$, we track the birth and death moment of connected components and cycles, thereby obtaining the Extended Persistence Diagram (EPD). The resulting diagram is represented as a multiset $\mathcal{D}_m$ in $\mathbb{R}^2$, defined as follows:
\begin{equation}
    \mathcal{D}_m = \operatorname{EPH}(\mathbb{F}_m) = \bigl\{\mu_m^k = (b_m^k, d_m^k) \mid k=1 \dots K \bigr\},
\end{equation}
where each point $\mu_m^k \in \mathbb{R}^2$ represents a topological feature that is born at threshold $b_m^k$ and persists until $d_m^k$ for modality $m \in \mathcal{M}$. Although the EPD fully characterizes the topological evolution induced by the filtration sequence in the form of a strict birth–death point set, it remains an unordered and variable-length multiset that is sensitive to minor perturbations caused by noise. Also. A multiset could not be directly amenable to numerical computation. These properties hinder the direct integration of EPD into end-to-end deep learning models. Therefore, we further parameterize the EPD into a fixed-dimensional Extended Persistence Image (EPI). This process begins by assigning a weight to each feature $\mu_k$ to emphasize robust structures with long lifespans over transient noise, using a weighting function $w(\mu_k)$ defined as $w(\mu_m^k) = |d_m^k - b_m^k|^\theta$, where $\theta$ denotes a scaling factor. 

Then, we map the weighted points onto a discretized grid $\Omega$ using Gaussian kernels. The value of the EPI at pixel $p$, centered at $\mathbf{z}_p$, is computed by summing the contributions of all features in the discrete persistence diagram $\mathcal{D}_m$ for modality $m \in \mathcal{M}$, and integrating this surface over each pixel grid $\Omega$, which is formulated as follows:
\begin{equation}
    \operatorname{EPI}_m(p) = \iint_{\Omega} \sum_{\mu_m^k \in \mathcal{D}_m} \frac{w(\mu_m^k)}{2\pi\sigma^2} \exp\left( -\frac{\| \mathbf{z}_p - \mu_m^k \|_2^2}{2\sigma^2} \right) d\mathbf{z}.
    \label{eq:epi}
\end{equation}

The resulting $\operatorname{EPI}$ effectively serves as the topological embedding for the following fusion module.

\subsection{Multi-view Geometric Fusion Network}
To effectively learn from both spatial and GEX data, while capturing cross-modality biological interdependencies, we construct separate views for each data modality. The resulting views are fused using a multi-view GCN network, which systematically integrates spatial organization, GEX profiles, and topological information. 

\noindent\textbf{Intra-Modality Graph Encoding.} To incorporate both spatial information and GEX patterns into the proposed model, we apply graph convolutions in parallel to the symmetrically normalized Laplacian matrix $\mathcal{T}$ of the expression-similarity adjacency matrix and the spatial adjacency matrix; we formalize the operation as follows:
\begin{equation}
	\begin{aligned}
		\mathbf{Z}_m^{(l+1)}
		&= \mathrm{ReLU}\Bigl(
		\mathcal{T}_m
		\,\mathbf{Z}_m^{(l)}\,
		\mathbf{\Theta}_m^{(l)}
		\Bigr), \ \  \forall m \in \mathcal{M},
	\end{aligned}
\end{equation}
where $\mathbf{Z}_{m}^{(l)}$ denotes the output embedding of the $l$-th GCN layer applied to the GEX graph or spatial graph. We define $\widetilde{\mathbf{A}}_{m}=\mathbf{A}_{m}+\mathbf{I}$ as the expression or spatial similarity adjacency matrix with self-loops, while $\widetilde{\mathbf{D}}_{m}$ denotes its diagonal degree matrix, thus, the symmetrically normalized Laplacian matrix $\mathcal{T}_m = \tilde{D}_m^{-\frac{1}{2}}\tilde{A}_m\tilde{D}_m^{-\frac{1}{2}}$. $\mathbf{\Theta}_{m}$ denotes the learnable weight matrix of the convolution layer for modality $m$.

\noindent\textbf{Cross-Modality Alignment \& Joint Learning.}
Motivated by the inherent correlation between gene-expression patterns and tissue architecture, we model both modality-specific characteristics and their shared latent representations. To this end, we employ a GCN with shared parameters to learn joint embeddings for the two modalities. We formalize this with the following propagation rule:
\begin{equation}
	\begin{aligned}
		\mathbf{Z}_{co,\,m}^{(l+1)}
		&= \mathrm{ReLU}\Bigl(
		\mathcal{T}_m
        \,\mathbf{Z}_{co,\,m}^{(l)}
		\,\mathbf{\Theta}_{\text{co}}^{(l)}
		\Bigr), \ \ \forall m \in \mathcal{M},
	\end{aligned}
\end{equation}
where $\mathbf{Z}_{co,\,m}^{(l+1)}$ denote the output co-spatial or co-expression embeddings of the $l$-th co-convolution layer, respectively. $\mathbf{\Theta}_{\mathrm{co}}$ denotes the shared learnable weight matrix of the co-convolution layer. With these two co-convolution embeddings for spatial locations and GEX profiles, we compute the final consensus co-embedding $\mathbf{Z}_{\mathrm{co}}$ using an aggregation function $\mathcal{F}_{agg}$ as follows:
\begin{equation}
	\mathbf{Z}_{co}^{(l)} =
    \mathcal{F}_{agg} \Bigr( \{ \mathbf{Z}_{co, m}^{(l)} \}_{m \in \mathcal{M}} \Bigr) = \frac{1}{|\mathcal{M}|} \sum_{m \in \mathcal{M}} \mathbf{Z}_{co,\, m}^{(l)}.
\end{equation}

To ensure that the co-embedding captures biological patterns shared across both modalities and that the latent representations progressively align during optimization, we introduce a consistency loss that minimizes the divergence between different modality, formalized as follows:
\begin{equation}
	\mathcal{L}_{\mathrm{con}}
	=
	\sum_{m, n \in \mathcal{M}, m \neq n} \bigl\lVert
	\widetilde{\mathbf{Z}}_{co,\, m}\,\widetilde{\mathbf{Z}}_{co,\, m}^{T}
	-
	\widetilde{\mathbf{Z}}_{co,\, n}\,\widetilde{\mathbf{Z}}_{co,\, n}^{T}
	\bigr\rVert_{2}^{2},
\end{equation}
where $\widetilde{\mathbf{Z}}_{co,\, m}$ and $\widetilde{\mathbf{Z}}_{co,\, n}$ denote the matrices obtained by normalizing $\mathbf{Z}_{co,\, m}$ and $\mathbf{Z}_{co,\, n}$, respectively.

\noindent\textbf{Topological Structure Embedding.}
The constructed EPI topological embedding is presented as an image and further processed using a two-dimensional convolutional layer. This layer extracts high-dimensional feature maps from the input EPI by applying a ReLU activation and a following max-pooling operation, thereby downsampling the features while preserving salient topological information. In brief, we propose the following extended persistence embedding (EPE) formulation:
\begin{equation}
	\begin{aligned}
		\operatorname{EPE}_{i,j}^{(m)}\! &=
		\operatorname{ReLU}\!\left(
		\sum_{p=0}^{k-1}\sum_{q=0}^{k-1}
		W^{(m)}_{p,q}\!
		\operatorname{EPI}^{(m)}_{i\ell + p,\;j\ell + q}
		+ b^{(m)}\!
		\right),
	\end{aligned}
	\label{eq:epe_conv}
\end{equation}
where $\mathrm{EPI}^{(m)}$ denote the EPI for each modality $m \in \mathcal{M}$. $\mathbf{W}^{(m)}$ and $\bm{b}^{(m)}$ denote the learnable weight matrices and bias vectors. While $k$ denotes the convolution kernel size, and $l$ denotes the stride length.

\noindent\textbf{Adaptive Attention Aggregation.}
To allocate the contributions of different modalities, we construct an attention-fusion layer that adaptively generates attention scores for each modality. Specifically, the attention score for each modality's EPE is computed as follows:
\begin{equation}
	\begin{aligned}
		\bm{a}_{E_m}
		= \operatorname{softmax}\Bigl(
		\mathbf{W}_2\,
		\sigma\bigl(\mathbf{W}_1\,\operatorname{EPE}^{(m)} + \bm{b}_1\bigr)
		\Bigr),
	\end{aligned}
\end{equation}
where $\mathbf{W}_1$ and $\mathbf{W}_2$ denote learnable feature transformation matrices, respectively, and $\bm{b}_1$ is the bias vector. Accordingly, we assign attention weights $a_{Em}$ to the EPE for modality $m \in \mathcal{M}$, attention weights $a_{n}$ to the GCN embedding for modality $n \in \mathcal{M}$, and $a_{co}$ to the spatial-expression co-embedding. We then aggregate the above embeddings using the computed attention weights to produce the final representations as follows:
\begin{equation}
	\begin{aligned}
		\mathbf{Z} 
		= \operatorname{MLP}\Bigl(
		\!\sum_{m\in \mathcal{M}}\bm{a}_{Em}\cdot \operatorname{EPE}^{(m)}
		+\!\sum_{n\in \mathcal{M}}a_{n}\,\mathbf{Z}_{n}
		+ a_{co}\,\mathbf{Z}_{co}\!
		\Bigr).
	\end{aligned}
\end{equation}

\subsection{Dual-Regularized Optimization}
Spatial transcriptomics data naturally contain spatial location information and exhibit a sparse, zero-inflated distribution. To leverage these two critical properties, we developed double design-optimization modules.

\noindent\textbf{Spatial-Coherence Constraint.}
To enhance the model's ability to learn from spatial locations, we propose a spatial-constraint module that explicitly preserves local neighborhood relationships in the embedding space. In brief, it operates via a two-part mechanism: (i) it increases the similarity between the embedding of cell $i$ and those of its spatial neighbors ($j\in\mathcal{N}_i$), and (ii) it decreases the similarity with non-neighboring cells for both the co-embedding and the view-specific embeddings. This strategy is formalized as follows:
\begin{equation}
  \label{eq:sco}
  \resizebox{\linewidth}{!}{$
  \begin{aligned}
\mathcal{L}_{sco} = - \sum_{i=1}^N \sum_{j \in \mathcal{N}_i} \Big[ y_{ij} \log \big( \sigma(S_{ij}) \cdot \Phi_{ii}^{+} \big) + (1 - y_{ij}) \log \big( (1 - \sigma(S_{ij})) \cdot \Phi_{ij}^{-} \big) \Big],
  \end{aligned} 
  $}
\end{equation}

where $y_{ij} = 1$ if $j\in\mathcal{N}_i$. And $S_{ij} = C(\mathbf{Z}_{i},\mathbf{Z}_{j})$ is the cosine similarity between two vectors, while $\sigma$ denotes the hyperbolic tanh activation function. View-specific embedding $\Phi_{ii}^{+} = \sum_{v=1}^V \exp(C(\mathbf{Z}_i, \mathbf{Z}_i^v))$ and $\Phi_{ij}^{-} = \sum_{v=1}^V \exp \big( (1 - S_{ij}) \cdot C(\mathbf{Z}_i, \mathbf{Z}_j^v) \big)$. $N$ denotes the number of cells in the spatial slice, and $V$ denotes the number of views.

\noindent\textbf{ZINB Probabilistic Reconstruction.}
Spatial transcriptomics data exhibit characteristic statistical properties: discrete counts with overdispersion and high sparsity. These properties can be modeled by a zero-inflated negative binomial (ZINB) distribution, whose mathematical form is given below:
\begin{align}
    \mathrm{NB}(X = x\! \mid\! \mu, \theta) 
        \! =\! \frac{\Gamma(x + \theta)}{x! \, \Gamma(\theta)}\! 
        \left( \frac{\theta}{\theta + \mu} \right)^\theta \!
        \left( \frac{\mu}{\theta + \mu} \right)^x, 
\end{align}

\begin{align}
	\mathrm{ZINB}(X = x\! \mid\! \pi, \mu, \theta)\! =\! \pi \operatorname{I}(x)\! +\! (1 - \pi) \, \mathrm{NB}(x),
\end{align}
where $\mu$ and $\theta$ denote the mean and dispersion of gene expression, while $\pi$ denotes the probability of zero inflation, $\operatorname{I}(\cdot)$ denote the indicator function. To incorporate these characteristics into spatial domain identification, we employ a three-layer MLP $f(\cdot)$ to estimate the zero-inflation probability matrix $\boldsymbol{\Pi}$, the mean matrix $\mathbf{M}$, and the dispersion matrix $\boldsymbol{\Theta}$, as follows:
\begin{equation}
	\begin{aligned}
		\bm{\Pi}
		=&\ \mathrm{sigmoid}\bigl(\mathbf{W}_{\pi}\,f(\mathbf{Z})\bigr);\\
		\mathbf{M}
		&= \exp\bigl(\mathbf{W}_{\mu}\,f(\mathbf{Z})\bigr);\\
		\bm{\Theta}
		&= \exp\bigl(\mathbf{W}_{\theta}\,f(\mathbf{Z})\bigr),
	\end{aligned}
\end{equation}
where $\mathbf{W}{\pi}$, $\mathbf{W}{\mu}$, and $\mathbf{W}_{\theta}$ denote the learnable weight matrices for $\boldsymbol{\Pi}$, $\mathbf{M}$, and $\boldsymbol{\Theta}$, respectively. To learn the model parameters, we minimize the negative log-likelihood of the ZINB model with respect to the observed $X$, which can be defined as follows:
\begin{equation}
	\mathcal{L}_{\mathrm{rec}}
	= -\sum
	\log(\mathrm{ZINB}\bigl(
	X
	\mid
	\pi,\mu,\theta
	\bigr)).
\end{equation}
\textbf{Total Objective.}
Ultimately, the fused embedding is jointly optimized using the consistency loss, the ZINB loss, and the spatial-constraint loss. We minimize the following total loss:
\begin{align}
	\mathcal{L}_{f} = \mathcal{L}_{rec} + \lambda_1\mathcal{L}_{con} + \lambda_2\mathcal{L}_{sco},
\end{align}
where $\lambda_1$ and $\lambda_2$ denote the trade-off hyperparameters for respective loss fuction terms.

\begin{table*}[t]\small
\centering
\scriptsize{
\resizebox{\linewidth}{!}{
\setlength\tabcolsep{3.pt}
\renewcommand\arraystretch{1.1}
\begin{tabular}{r||ccccccccc}
\hline\thickhline

\cellcolor{gray!20} & \multicolumn{9}{c}{\cellcolor{gray!20}\textbf{ARI}} \\
\cline{2-10}

\multirow{-2}{*}{\cellcolor{gray!20}Datasets} 
& \cellcolor{gray!20}SCANPY'18 & \cellcolor{gray!20}SpaGCN'21 & \cellcolor{gray!20}DeepST'22 & \cellcolor{gray!20}stLearn'23 & \cellcolor{gray!20}SCGDL'23 & \cellcolor{gray!20}GraphST'23 &  \cellcolor{gray!20}Spatial-MGCN'23 & \cellcolor{gray!20}STAIG'25 & \cellcolor{gray!20}\textbf{SPHENIC} \\
\hline\hline

HBC & $48.52_{\pm 0.14}$ & $56.79_{\pm 2.81}$ & $56.30_{\pm 0.89}$ & $55.02_{\pm 3.36}$ & $32.88_{\pm 2.51}$ & $51.83_{\pm 0.57}$ & $\underline{64.04}_{\pm 3.24}$ & $50.69_{\pm 2.65}$ & \cellcolor[HTML]{D7F6FF}$\bm{68.23}_{\pm 2.36}$
\\

\rowcolor{gray!10} MBA & $\underline{39.65}_{\pm 0.95}$ & $34.42_{\pm 0.89}$ & $25.91_{\pm 3.43}$ & $34.13_{\pm 0.19}$ & $25.81_{\pm 1.23}$ & $29.33_{\pm 0.13}$ & $39.58_{\pm 0.02}$ & $37.01_{\pm 0.71}$ & \cellcolor[HTML]{D7F6FF}$\bm{48.79}_{\pm 1.68}$
\\

151507 & $25.53_{\pm 1.23}$ & $38.26_{\pm 3.05}$ & $52.24_{\pm 2.76}$ & $42.95_{\pm 1.04}$ & $46.26_{\pm 2.41}$ & $44.79_{\pm 0.30}$ & $\underline{62.68}_{\pm 4.60}$ & $54.22_{\pm 1.68}$ & \cellcolor[HTML]{D7F6FF}$\bm{65.33}_{\pm 1.74}$
\\

\rowcolor{gray!10} 151508 & $14.25_{\pm 0.70}$ & $39.48_{\pm 2.15}$ & $42.89_{\pm 5.77}$ & $40.03_{\pm 0.68}$ & $35.54_{\pm 1.82}$ & $35.18_{\pm 1.01}$ & $\underline{50.66}_{\pm 1.04}$ & $47.69_{\pm 3.98}$ & \cellcolor[HTML]{D7F6FF}$\bm{53.54}_{\pm 4.89}$
\\

151509 & $14.31_{\pm 1.17}$ & $34.08_{\pm 2.39}$ & $47.60_{\pm 6.47}$ & $29.52_{\pm 0.75}$ & $37.37_{\pm 3.40}$ & $36.71_{\pm 5.81}$ & $\underline{58.10}_{\pm 3.87}$ & $55.02_{\pm 0.46}$ & \cellcolor[HTML]{D7F6FF}$\bm{60.51}_{\pm 6.14}$
\\

\rowcolor{gray!10} 151510 & $17.04_{\pm 0.52}$ & $38.36_{\pm 3.26}$ & $46.57_{\pm 2.11}$ & $34.09_{\pm 4.33}$ & $30.57_{\pm 0.34}$ & $39.73_{\pm 4.05}$ & $ \bm{57.06}_{\pm 3.64}$ & $51.26_{\pm 0.21}$ & \cellcolor[HTML]{D7F6FF}$\underline{53.97}_{\pm 3.41}$ 
\\

151669 & $25.50_{\pm 2.86}$ & $26.70_{\pm 4.79}$ & $27.73_{\pm 1.81}$ & $25.23_{\pm 0.19}$ & $25.15_{\pm 0.84}$ & $29.98_{\pm 5.75}$ & $\underline{45.14}_{\pm 5.91}$ & $33.40_{\pm 2.65}$ & \cellcolor[HTML]{D7F6FF}$\bm{49.04}_{\pm 5.35}$
\\

\rowcolor{gray!10} 151670 & $24.34_{\pm 1.61}$ & $23.80_{\pm 2.88}$ & $28.56_{\pm 2.76}$ & $23.46_{\pm 0.60}$ & $20.94_{\pm 1.59}$ & $25.97_{\pm 1.45}$ & $\bm{42.67}_{\pm 5.79}$ & $24.62_{\pm 2.96}$ & \cellcolor[HTML]{D7F6FF}$\underline{42.60}_{\pm 5.61}$
\\

151671 & $21.97_{\pm 1.58}$ & $40.76_{\pm 1.08}$ & $37.24_{\pm 3.75}$ & $33.22_{\pm 4.78}$ & $23.43_{\pm 0.26}$ & $38.83_{\pm 0.95}$ & $\bm{65.75}_{\pm 0.61}$ & $48.87_{\pm 5.95}$ & \cellcolor[HTML]{D7F6FF}$\underline{65.14}_{\pm 6.40}$
\\

\rowcolor{gray!10} 151673 & $17.46_{\pm 1.90}$ & $44.73_{\pm 1.04}$ & $52.82_{\pm 0.54}$ & $35.34_{\pm 2.69}$ & $32.48_{\pm 1.03}$ & $48.00_{\pm 2.28}$ & $\underline{56.92}_{\pm 4.93}$ & $51.88_{\pm 5.73}$ & \cellcolor[HTML]{D7F6FF}$\bm{58.17}_{\pm 2.14}$ \\

151674 & $23.95_{\pm 1.97}$ & $33.26_{\pm 2.97}$ & $54.51_{\pm 3.19}$ & $30.60_{\pm 2.76}$ & $23.41_{\pm 0.78}$ & $44.03_{\pm 3.65}$ & $\underline{56.10}_{\pm 1.71}$ & $56.03_{\pm 3.84}$ & \cellcolor[HTML]{D7F6FF}$\bm{56.41}_{\pm 2.86}$ \\

\hline
\rowcolor{gray!10} \textit{Avg} & $24.78_{\pm 10.62}$ & $37.33_{\pm 8.85}$ & $42.94_{\pm 11.37}$ & $34.87_{\pm 8.79}$ & $30.35_{\pm 7.57}$ & $38.58_{\pm 8.17}$ & $\underline{54.43}_{\pm 8.82}$ & $46.43_{\pm 10.20}$ & \cellcolor[HTML]{D7F6FF}$\bm{56.52}_{\pm 7.95}^{\textcolor{RedOrange}{\ \uparrow 2.09}}$\\

\hline \hline

\cellcolor{gray!20}Datasets\  & \multicolumn{9}{c}{\cellcolor{gray!20}\textbf{NMI}} \\

\hline\hline

HBC & $60.60_{\pm 0.21}$ & $65.71_{\pm 0.59}$ & $69.33_{\pm 0.58}$ & $66.47_{\pm 1.84}$ & $48.28_{\pm 5.68}$ & $65.86_{\pm 0.39}$ & $\underline{69.41}_{\pm 1.88}$ & $68.08_{\pm 1.54}$ & \cellcolor[HTML]{D7F6FF}$\bm{69.54}_{\pm 1.88}$ \\

\rowcolor{gray!10} MBA & $63.89_{\pm 0.19}$ & $64.88_{\pm 1.51}$ & $55.23_{\pm 2.75}$ & $64.55_{\pm 0.04}$ & $62.04_{\pm 0.36}$ & $67.69_{\pm 0.15}$ & $63.65_{\pm 1.99}$ & $\underline{67.75}_{\pm 0.14}$ & \cellcolor[HTML]{D7F6FF}$\bm{69.09}_{\pm 1.65}$ \\

151507 & $34.36_{\pm 0.67}$ & $49.61_{\pm 3.40}$ & $64.54_{\pm 3.99}$ & $54.35_{\pm 0.55}$ & $52.96_{\pm 1.26}$ & $57.40_{\pm 0.69}$ & $\underline{70.92}_{\pm 1.03}$ & $68.49_{\pm 2.00}$ & \cellcolor[HTML]{D7F6FF}$\bm{74.08}_{\pm 2.13}$ \\

\rowcolor{gray!10} 151508 & $19.54_{\pm 1.03}$ & $51.04_{\pm 1.35}$ & $56.81_{\pm 4.56}$ & $51.57_{\pm 0.76}$ & $45.61_{\pm 0.16}$ & $52.57_{\pm 0.60}$ &  $60.42_{\pm 2.38}$ & $\underline{61.93}_{\pm 2.18}$ & \cellcolor[HTML]{D7F6FF}$\bm{63.12}_{\pm 1.16}$ \\

151509 & $26.00_{\pm 0.51}$ & $51.35_{\pm 3.52}$ & $62.92_{\pm 1.87}$ & $49.01_{\pm 0.96}$ & $51.34_{\pm 1.18}$ & $53.74_{\pm 1.58}$ & $67.08_{\pm 2.28}$ & $\underline{67.83}_{\pm 0.21}$ & \cellcolor[HTML]{D7F6FF}$\bm{69.61}_{\pm 0.27}$ \\

\rowcolor{gray!10} 151510 & $26.29_{\pm 1.93}$ & $49.88_{\pm 3.84}$ & $61.26_{\pm 1.08}$ & $47.77_{\pm 2.03}$ & $47.34_{\pm 0.42}$ & $54.70_{\pm 0.70}$ & $65.39_{\pm 2.72}$ & $\underline{65.44}_{\pm 1.25}$ & \cellcolor[HTML]{D7F6FF}$\bm{65.63}_{\pm 0.88}$ \\

151669 & $26.79_{\pm 1.08}$ & $45.41_{\pm 6.29}$ & $52.16_{\pm 0.93}$ & $45.18_{\pm 0.67}$ & $41.26_{\pm 0.61}$ & $51.02_{\pm 2.71}$ & $\underline{59.48}_{\pm 6.34}$ & $49.48_{\pm 1.88}$ & \cellcolor[HTML]{D7F6FF}$\bm{61.84}_{\pm 0.48}$ \\

\rowcolor{gray!10} 151670 & $24.73_{\pm 4.05}$ & $42.24_{\pm 1.90}$ & $51.65_{\pm 2.08}$ & $43.79_{\pm 1.22}$ & $36.75_{\pm 1.30}$ & $45.76_{\pm 1.64}$ & $\underline{55.13}_{\pm 2.60}$ & $48.86_{\pm 1.29}$ & \cellcolor[HTML]{D7F6FF}$\bm{60.23}_{\pm 1.52}$ \\

151671 & $23.34_{\pm 2.53}$ & $55.50_{\pm 1.48}$ & $58.84_{\pm 3.32}$ & $50.76_{\pm 1.53}$ & $40.04_{\pm 0.45}$ & $57.27_{\pm 0.77}$ & $63.41_{\pm 0.42}$ & $\underline{66.78}_{\pm 2.72}$ & \cellcolor[HTML]{D7F6FF}$\bm{73.50}_{\pm 2.19}$ \\

\rowcolor{gray!10} 151673 & $29.44_{\pm 1.73}$ & $59.08_{\pm 1.20}$ & $\underline{67.71}_{\pm 0.66}$ & $51.92_{\pm 0.76}$ & $43.04_{\pm 0.46}$ & $62.54_{\pm 1.46}$ & $67.08_{\pm 2.14}$ & $67.36_{\pm 2.64}$ & \cellcolor[HTML]{D7F6FF}$\bm{68.63}_{\pm 0.94}$ \\

151674 & $30.66_{\pm 1.39}$ & $45.86_{\pm 1.48}$ & $64.31_{\pm 2.88}$ & $49.52_{\pm 2.22}$ & $36.33_{\pm 0.93}$ & $55.32_{\pm 2.48}$ & $64.94_{\pm 1.93}$ & $\underline{66.97}_{\pm 1.86}$ & \cellcolor[HTML]{D7F6FF}$\bm{65.56}_{\pm 1.21}$ \\

\hline
\rowcolor{gray!10} \textit{Avg} & $33.24_{\pm 14.87}$ & $52.78_{\pm 7.73}$ & $60.43_{\pm 5.99}$ & $52.26_{\pm 7.21}$ & $45.91_{\pm 7.66}$ & $56.72_{\pm 6.51}$ & $\underline{64.26}_{\pm 4.59}$ & $63.54_{\pm 7.33}$ & \cellcolor[HTML]{D7F6FF}$\bm{67.30}_{\pm 4.48}^{\textcolor{RedOrange}{\ \uparrow 3.04}}$ \\
\hline

\end{tabular}}}
\vspace{-5pt}
\captionsetup{font=small}
\caption{{
Comparison with state-of-the-art methods on the eleven distinct spatial transcriptomic slices. The optimal and second-best results are denoted by \textbf{boldface} and \underline{underlining}, respectively. \redup{} means improved performance compared with the second-best baseline.
}}
\label{tab:performance}
\vspace{-10pt}
\end{table*}

\section{Experiments}

\subsection{Experiment Settings}
\noindent\textbf{Benchmark Datasets and Evaluation.}
The proposed SPHENIC model was evaluated on three publicly available spatial transcriptomics datasets, DLPFC (comprising 9 distinct tissue sections), 10x Visium human breast cancer dataset (HBC), and mouse brain anterior dataset (MBA) \cite{maynard2021DLPFC,buache2011HBC}. Clustering performance was assessed using the adjusted Rand index (ARI) and normalized mutual information (NMI).

\noindent\textbf{Compared Baseline Methods.} 
To enable a fair comparison, we benchmark our model against eight current representative baseline methods: SCANPY \cite{wolf2018scanpy}, SpaGCN \cite{hu2021spagcn}, DeepST \cite{xu2022deepst}, stLearn \cite{pham2023robust}, SCGDL \cite{liu2023scgdl},  GraphST \cite{long2023graphst}, Spatial-MGCN \cite{wang2023spatial-mgcn} and STAIG \cite{yang2025staig}.

\noindent\textbf{Implementation Details.} 
All experiments were conducted on a single NVIDIA virtual GPU (vGPU) rented from AutoDL, providing up to 48 GB of memory and a 25-core Intel Xeon Platinum 8481C processor. The model was implemented in PyTorch 2.1.0 and executed with CUDA 12.1. For each dataset, training was terminated after 100 epochs. All baseline methods were run with the default hyperparameters recommended in their original publications.

\subsection{Clustering Performance}
\begin{figure*}[t]
	\centering
	\includegraphics[width=0.95\textwidth]{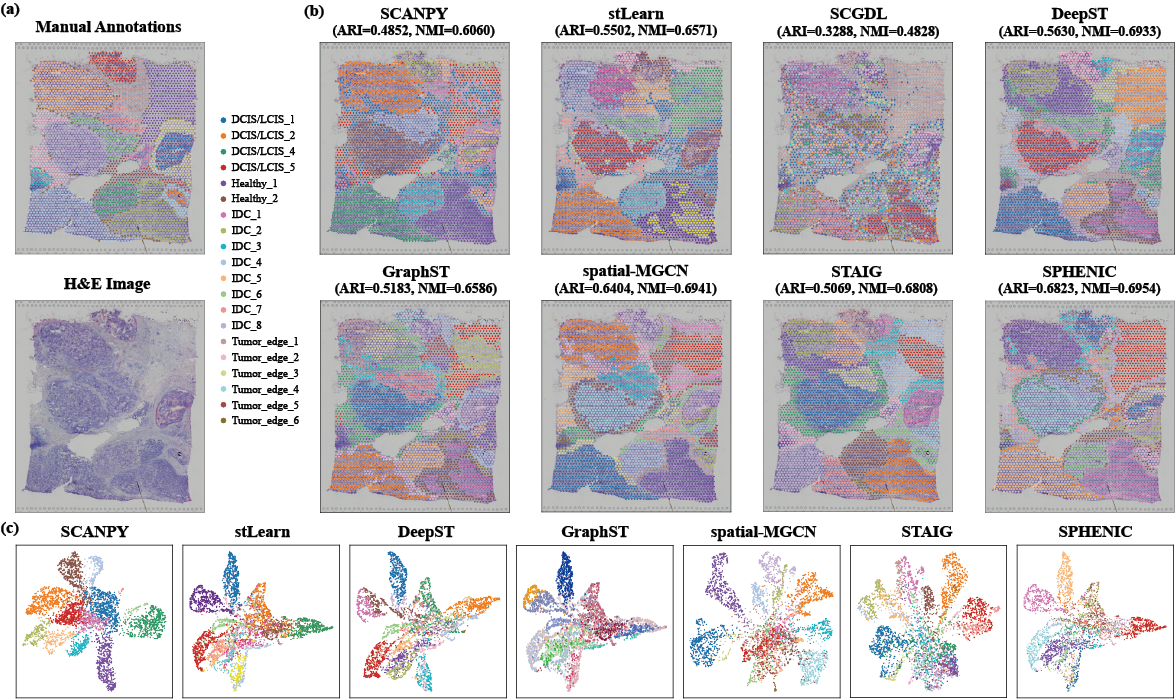}
	\caption{Comparison experiments on HBC dataset between SPHENIC and other baselines. (a) Manual ground-truth annotations and histological image of the HBC dataset. (b) Spatial clustering result visualization by SCANPY, stLearn, SCGDL, DeepST, GraphST, spatial-MGCN, STAIG, and SPHENIC. (c) UMAP visualization of detected clusters by the above baselines.}
	\label{fig_location}
\end{figure*}

The comparative performance of nine methods is summarized in Table \ref{tab:performance}. Fig. \ref{fig_location} provides a visual comparison of the clustering outputs produced by each method. The above results reveal that:

(i) SPHENIC delivers state-of-the-art clustering accuracy in architecturally complex tissues. With respect to the ARI term, SPHENIC exceeds the next-best method (Spatial-MGCN) by 4.19\% on the HBC dataset and 9.14\% on the mouse-anterior dataset compared to the second-best method (SCANPY), underscoring its strength in identifying domains within challenging tissue architectures.

(ii) In spatially simpler structures such as the DLPFC slices, SPHENIC surpasses the next-best method (Spatial-MGCN for slice 151507, 151508, and 151569) by 2.65\%, 2.88\%, and 3.90\% ARI, respectively. These findings confirm SPHENIC’s generalizability and its ability to extract fundamental biological signals irrespective of spatial complexity.

(iii)SPHENIC accurately delineates tumor boundaries in the HBC dataset, as illustrated in the Fig. \ref{fig_location}. Our SPHENIC shows strong agreement with manual annotations in identifying key tumor-microenvironment components (e.g., 'IDC\_4', 'healthy\_1').  Unlike baselines that yield fragmented clusters and tumor-immune misclassifications, SPHENIC preserves biologically valid structures, achieving expert-level spatial recognition in carcinomas.

Additionally, we performed scalability analysis of all the baselines in the HBC dataset, the 151507 dataset, and the MBA dataset, which are demonstrated in the supplementary materials.

\begin{table}[t]\small
\centering
\newcommand{\cc}{\cellcolor[HTML]{D7F6FF}}
\resizebox{0.95\columnwidth}{!}{
\setlength\tabcolsep{4pt}
\renewcommand\arraystretch{1.0}
\begin{tabular}{ccc || cc}
\hline \thickhline
\rowcolor{lightgray}
& \multicolumn{2}{c||}{Module} & \multicolumn{2}{c}{Metric} \\
\rowcolor{lightgray}
\multirow{-2}{*}{Dataset} & Topo & DualRO & \textit{ARI} & \textit{NMI} \\
\hline \hline

\multirow{4}{*}{\textbf{HBC}} &
\multicolumn{2}{c||}{baseline} & $52.89_{\pm 2.38}$ & $62.61_{\pm 2.14}$ \\
\cdashline{2-5} 
& \ding{51} & & $57.99_{\pm 2.73}$ & $63.93_{\pm 2.54}$ \\
& & \ding{51} & $63.39_{\pm 4.23}$ & $64.94_{\pm 2.71}$ \\

& \cc \ding{51} & \cc \ding{51} & \cc $\textbf{68.23}_{\pm 2.36}$ & \cc $\textbf{69.54}_{\pm 1.88}$ \\ 
\hline \hline
\multirow{4}{*}{\textbf{151507}} 
& \multicolumn{2}{c||}{baseline}& $53.22_{\pm 1.37}$ & $63.79_{\pm 1.33}$ \\
\cdashline{2-5}
& \ding{51} &  & $59.92_{\pm 3.08}$ & $66.95_{\pm 2.06}$ \\
&  & \ding{51} & $62.03_{\pm 2.65}$ & $71.78_{\pm 1.70}$ \\
& \cc \ding{51} & \cc \ding{51} & \cc $\textbf{65.33}_{\pm 1.74}$ & \cc $\textbf{74.08}_{\pm 2.13}$ \\
\hline
\end{tabular}}
\caption{
Ablative study of key modules in our SPHENIC on HBC and 151507 datasets. This table demonstrates the effects of the EPH module and Dual-Regularized Optimization over the baselines, which remove these modules respectively or simultaneously. 'Topo' means EPH module.
}
\label{tab:abla}
\vspace{-10pt}
\end{table}

\subsection{Ablation Study}
To demonstrate the effectiveness of the proposed modules, we construct several variants of SPHENIC: (i) Omits the invariant topological feature-extraction module; (ii) Excludes the dual-regularized optimization module; and (iii) Removes both modules. As shown in the Table \ref{tab:abla}, in the HBC dataset, omitting the topological module (‘w/o Topo’) reduces ARI by 4.84 \% and 3.30 \% relative to the full model, thereby validating the contribution of the topological learning component. Similarly, removing the dual-regularized optimization module (‘w/o DualRO’) decreases ARI by 10.24 \% and 5.61 \%, confirming the importance of explicitly modeling the spatial

\begin{figure}[H]
	\centering
	\includegraphics[width=0.95\columnwidth]{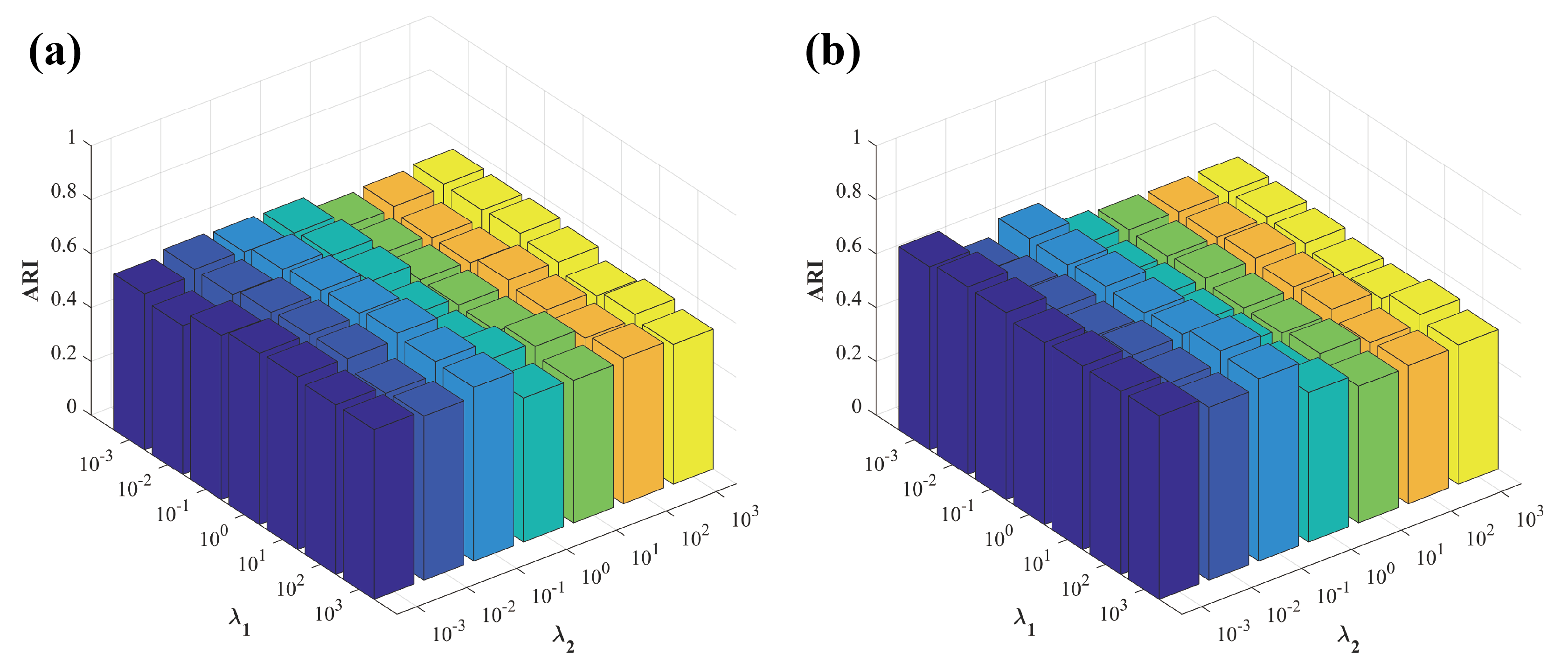} 
	\caption{Hyperparameter sensitivity analysis on HBC dataset and DLPFC-151507, respectively, in ARI term.}
	\label{fig_para}
\end{figure}
\noindent neighbourhoods and non-neighbourhoods. We further evaluate a variant lacking both modules. The complete SPHENIC model improves ARI by 15.34 \% and 6.93 \% over this variant, demonstrating the synergistic benefit of combining topological and spatial-constraint learning. Ablation experiment on the MBA dataset is demonstrated in the supplementary materials.

\subsection{Model Analysis}
\noindent\textbf{Hyperparameter Sensitivity Analysis.}
We evaluate the effect of different hyperparameters on the performance of our proposed SPHENIC in the HBC dataset, which includes the trade-off coefficients of the total loss function ($\lambda_1$ and $\lambda_2$). Fig. \ref{fig_para} demonstrates the ARI term of our SPHENIC when $\lambda_1$ and $\lambda_2$ are varied from $10^{-3}$ to $10^3$. From this figure, the proposed SPHENIC is insensitive to both $\lambda_1$ and $\lambda_2$ in the range $10^{-3}$ to $10^{-1}$, further illustrating the robustness of our proposed SPHENIC. Sensitivity analysis of the EPI hyperparameters is further demonstrated in the supplementary materials.
\begin{figure}[t]
	\centering
	\includegraphics[width=1\columnwidth]{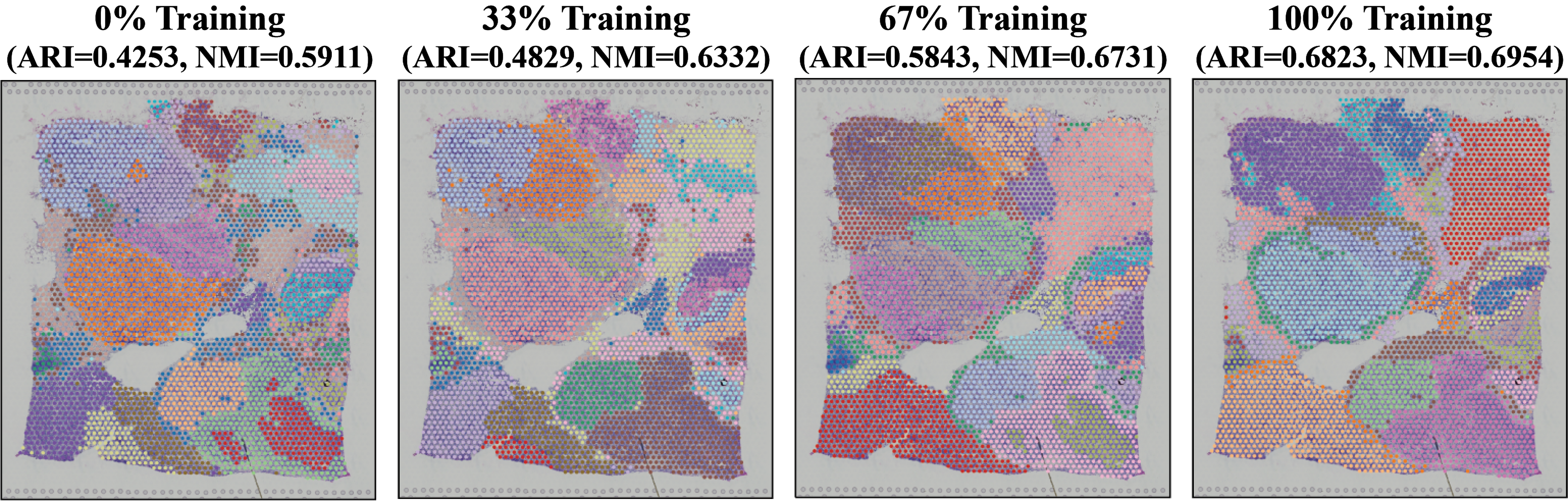} 
	\caption{Visualization of spatial clustering by SPHENIC model in different training stages}
	\label{fig_epoch}
	\vspace{-10pt}
\end{figure}

\noindent\textbf{Visualization Analysis.}
In addition, to further verify the effectiveness of our SPHENIC, we visualize the spatial clustering results of the proposed SPHENIC in different training stages ($0\%$, $33\%$, $67\%$ and $100\%$ of training epochs) by the UMAP visualization methods in Fig. \ref{fig_epoch} \cite{mcinnes2018umap}. We can clearly observe that the spatial clustering structure of our comprehensive multi-view fused embedding $\mathbf{Z}$ progressively refines across the training epochs, along with consistent improvement in both ARI and NMI terms.

\noindent\textbf{Gene Enhancement Analysis.}
To validate the proposed SPHENIC's dual-module optimization strategy in preserving spatial expression patterns while reducing the noise artifacts, we evaluated its performance on the HBC dataset by examining the spatial distributions of characteristic marker genes (\textit{DSP}, \textit{IFT122}, \textit{CCL21}, \textit{CTTN}, \textit{GNG5}, \textit{FARP1}). Fig. S1 in the supplementary materials shows that SPHENIC successfully reconstructs biologically plausible expression patterns while significantly reducing technical noise compared to raw data. The system maintains consistent spatial distributions of these markers that align with established biological knowledge, confirming its ability to (i) filter out non-informative noise, (ii) recover true biological signals from degraded data, and (iii) faithfully reconstruct spatial expression profiles.

\section{Conclusions}

In this paper, we present SPHENIC, a topology-aware clustering framework for spatial transcriptomics data. By leveraging extended persistent homology features, the proposed model learns more robust topological information and is less sensitive to intrinsic noise. Meanwhile, we leverage distribution constraints and spatial coordinates to supervise the embedding space through a novel DualRO module, forcing proximity between a cell’s embedding and those of its spatial neighbors while separating it from non-neighboring cells, thereby producing high-quality spatial embeddings for applications. Extensive validation experiments demonstrate the superiority of the proposed SPHENIC. Nevertheless, the principal limitation of our approach is the computational overhead associated with extended persistent homology, which may impede scalability in application scenarios.

\appendix



\bibliographystyle{named}
\bibliography{ijcai26}

\end{document}